\documentclass[11pt, a4paper, onecolumn]{article}

\usepackage[english]{babel}

\usepackage{geometry}     
\usepackage{natbib}
\usepackage{amsmath}
\usepackage{graphicx}
\usepackage{hyperref}
\hypersetup{
    colorlinks=true,   
    allcolors=black,
    urlcolor=blue,
    pdfborder={0 0 0},
}
\usepackage{authblk}
\usepackage{caption} 
\usepackage{times}

\title{Agentic Insight Generation in VSM Simulations}
\author[1]{Micha Selak}
\author[1]{Dirk Krechel}
\author[1]{Adrian Ulges}
\author[2]{\protect\\ Sven Spieckermann}
\author[2]{Niklas Stoehr}
\author[2]{Andreas Loehr}

\affil[1]{RheinMain University of Applied Sciences, Wiesbaden, GERMANY}
\affil[2]{SimPlan AG, Hanau, GERMANY}
\date{\vspace{-5ex}}

\begin{document}
\maketitle

\vspace{-12pt}

\section*{ABSTRACT}

Extracting actionable insights from complex value stream map simulations can be challenging, time-
consuming, and error-prone. Recent advances in large language models offer new avenues to support
users with this task. While existing approaches excel at processing raw data to gain information, they are structurally unfit to pick up on subtle situational differences needed to distinguish similar data sources in this domain. To address this issue, we propose a decoupled, two-step agentic architecture. By separating orchestration from data analysis, the system leverages progressive data discovery infused with domain expert knowledge. This architecture allows the orchestration to intelligently select data sources and perform multi-hop reasoning across data structures while maintaining a slim internal context. Results from multiple state-of-the-art large language models demonstrate the framework's viability: with top-tier models achieving accuracies of up to 86\% and demonstrating high robustness across evaluation runs.

\section*{Preprint Notice}
This is the author’s preprint of a paper intended for publication at the Winter Simulation Conference (WSC) 2026. 
The final authenticated version will be published by IEEE and will be available via IEEE Xplore.

\section{INTRODUCTION}
\label{sec:intro}

Simulation-based optimization serves as an iterative decision-support mechanism for improving production processes. Typically, it alternates between two steps: First, a  new production scenario is systematically created by varying relevant parameters such as resource allocation, process sequences, and scheduling policies. Second, the scenario is then evaluated using a simulation model that captures the dynamic behavior of the production system under realistic operating conditions. Simulation results are then analyzed to assess the effectiveness of the scenario. 
Based on this analysis, promising parameter settings are explored and reintroduced into subsequent simulation runs, forming a closed feedback loop. This iterative process enables the exploration of a large solution space and supports the identification of near-optimal production configurations without disrupting real-world operations.

Recently, the rise of Large Language Models (LLMs) -- which serve as general-purpose problem solvers across various domains such as software engineering~\citep{dong2025ASO}, personal assistance~\citep{li2024personal} or even research~\citep{yamada2025ai} --
has raised the question to which extent LLM-based successes can be achieved in the domain of process simulation. In theory, LLMs hold promise to support {\it both} steps of the above feedback loop -- scenario creation as well as simulation result interpretation. Correspondingly, LLM assistance in the simulation domain has appeared as a research field~\citep{11338856,10838994}. However, a key challenge lies in the fact that the more faithfully a simulation model represents reality, the more complex it becomes. This can lead to vast inputs, as well as complex reasoning and data wrangling challenges to the LLM.  

A promising paradigm to overcome these limitations is the deployment of LLM-based agents~\citep{YaoZYDSN023}: Instead of challenging the LLM with the monolithic task, data load and reasoning steps are chunked into more digestible bits. Specifically, the LLM-based agent operates in a loop, in which tools allow him to access certain aspects of the target problem and reason over them in multiple steps. 

In this paper, we present an agentic LLM-based approach that supports experts with simulation-based optimization. Our focus is on the second step of the simulation loop, i.e. simulation result interpretation: We investigate to what extent an LLM-based agent can answer pracitioners's questions about simulation results, given the simulation's output and the underlying simulation model. Thereby, we focus on {\it value stream mapping} (VSM) as a simulation domain:

Value streams encompass the complete sequence of actions, information flows, and material movements required to deliver a product or service. To optimize these processes, experts use VSM, a core lean management methodology~\citep{Rother2003}, to create holistic visualizations of this flow from start to finish. While traditional VSM provides a valuable but static baseline, integrating simulation transforms these maps into dynamic models capable of capturing complex, real-world variability. While recent research has explored deploying LLM agents in production optimization, prevailing approaches typically focus on code generation to calculate metrics from raw data~\citep{Keskin2025}. However, since our VSM simulations produce large quantities of pre-calculated Key Performance Indicators (KPIs) that are semantically similar, our work addresses this by integrating domain expert knowledge into the retrieval process to identify the correct data source.

To achieve this, we propose a suitable agentic design: An orchestration agent focuses on high-level reasoning and answer generation, utilizing (a) a sub-agent (or subworkflow) specialized in summarizing the simulation data, and (b) tools to access specific parts of the VSM.  This enables the orchestration agent to progressively discover the required information, without filling the context window with unnecessary data and thereby avoiding context rot~\citep{hong2025context}. To validate this approach, we introduce an evaluation methodology utilizing an LLM-as-a-Judge backed by a human validation baseline to ensure reliability. Furthermore, we provide empirical cross-model benchmarking, comparing a range of LLMs — from efficient open-source to state-of-the-art (SOTA) closed-source — within this decoupled framework.

\section{RELATED WORK}
\label{sec:related}
This section contextualizes our proposed architecture by reviewing foundational literature across the three intersecting domains that underpin our approach: the evolution of general question answering methodologies in recent decades, the deployment of LLM agents within operations research, and the current landscape of machine learning in VSM.

\subsection{Evolution of Question Answering}

Question-answering systems typically employ information retrieval (IR) components to formulate their responses. Over the past decade, IR has transitioned from rigid keyword matching to dynamic, reasoning-driven architectures. While standard single-pass Retrieval-Augmented Generation (RAG)\citep{lewis2020retrieval} significantly improved the ability to ground LLMs in external facts~\citep{shuster-etal-2021-retrieval-augmentation}, such pipelines inherently struggle with the complex, interconnected nature of highly structured data. Consequently, SOTA IR has largely moved towards agentic approaches~\citep{AsaiWWSH24}. Unlike basic retrieve-and-answer workflows, agentic systems utilize the LLM as an autonomous reasoning engine capable of breaking down complex queries, navigating tools, and reflecting on retrieved data across multiple steps. This multi-step autonomy makes agents well-suited for navigating datasets where standard single-pass RAG methods fail.

\subsection{LLM Agents in Operations Research and Simulation}

The powerful reasoning capabilities of LLM agents have prompted their application in operations research and simulation domains~\citep{li2023large,qiao2023taskweaver}. The prevailing paradigm in these systems is the use of coding agents that generate program code (e.g., python scripts). This code is then executed to query or process raw data. This approach excels in domains where metrics first need to be calculated from clearly structured data, which then are used to derive answers. In our case, however, analyzing single bits of raw data only plays a secondary role since the simulation data already contains the most of the required metrics or domain KPIs. The key challenge lies in distinguishing between the fast amount of similar KPIs based on situational changes, and performing multi-hop reasoning on different data structures to gain meaningful insights.
Other approaches from the simulation domain, such as simio~\citep{simio}, support simulation modeling. While these further demonstrate the capabilities of LLM based agents in the field, they are similarly not designed for our use case.

\subsection{Machine Learning in VSM}

Within the specific domain of VSM, data-driven approaches have traditionally relied on standard machine learning.~\citet{Wrzalik2023} developed random forest models to suggest VSM corrections, while \citet{Becker2020} and \citet{Vernickel2020} applied network embeddings and XGBoost models for material flow optimization. While effective for narrow, predefined tasks, these models lack natural language flexibility and direct user interaction. More recently, LLMs were used in the modeling of VSMs and very basic VSM understanding~\citep{selak2025llm}. While the approach simplified the generation of VSMs, extracting insights from them and their simulation outputs remains a distinct and highly complex challenge. Highlighting this difficulty,~\citet{Magnus2024} evaluated vanilla ChatGPT as a Lean Manufacturing consultant and found that, while promising, it produced unreliable results, underscoring the need for more constrained architectures. Thus, dynamically interpreting VSMs and their simulations in order to gain meaning full insights remains an open gap, that the decoupled agentic architecture presented in this paper explicitly addresses.

\section{APPROACH}
\label{sec:appro}

This section details the proposed methodology for enabling agentic driven insight generation within the VSM simulation environment. Section~\ref{subsec:vsm-sim-details} outlines the broad data structure present in VSM simulations. Next, Section~\ref{subsec:data} describes the curated dataset used to develop and evaluate the presented approach. Finally, in Section~\ref{subsec:architecture} we present the core technical contribution of this work.

\subsection{VSM Simulation Data Structure and Composition}
\label{subsec:vsm-sim-details}
\begin{figure}[!hb]
{
\centering
\includegraphics[width=0.70\textwidth]{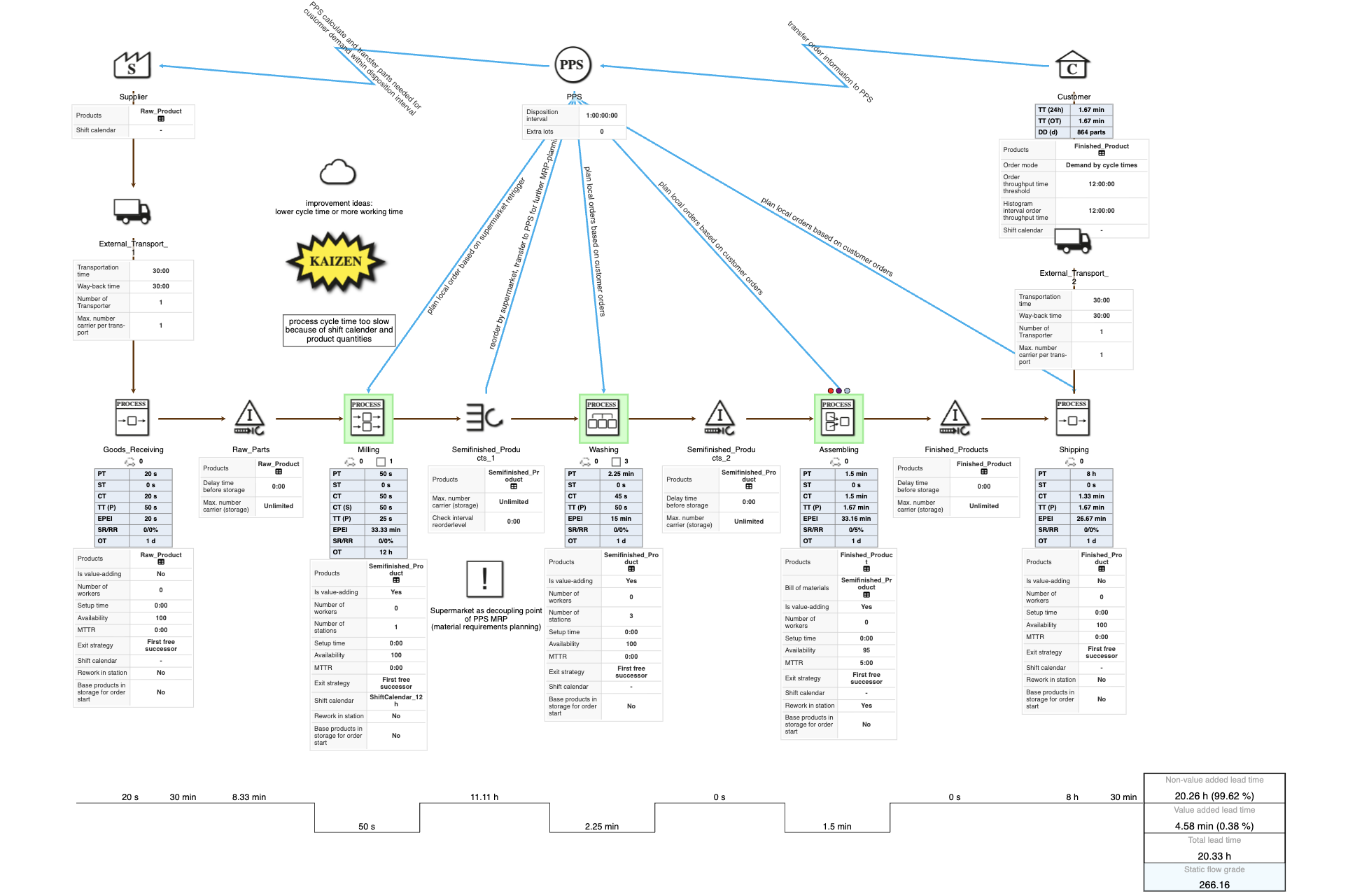}
\caption{A simple Example VSM.\label{fig:vsm}}
}
\end{figure}

VSM and their simulation data are used as the contextual base of questions and answers in this study. This context data is formatted in JSON and typically spans megabytes per instance. Figure~\ref{fig:vsm} depicts a simple digital VSM model example.

The VSM is modeled as an attributed graph and viewed primarily through a logistical lens. The scale of these graphs generally range from 10 to 30 nodes, though highly complex VSMs can scale over 100 nodes. The nodes represent distinct logistical entities, such as different kinds of processing stations, warehouses, customers and more. The edges connecting these nodes model the flow of either materials or information. These connections are governed by strict topological rules, meaning not all node types can be arbitrarily connected to one another. Furthermore, nodes encompass varying attributes tailored to their specific entity type. For example, some warehouse nodes can define parameters like maximum capacity limits or safety stock levels; process nodes can specify cycle times, repair durations, or worker allocations.

The VSM is simulated via Discrete Event Simulation. A single simulation run can model an operational time frame ranging from a few hours to several weeks. This execution generates a substantial volume of dynamic output. Alongside raw multi-variate time-series (i.e. stock levels of different products in a warehouse aggregated in different ways), various statistical metrics and specific VSM KPIs are calculated and included in the data. These data elements are systematically organized by subject matter into 11 high-level categories, separating distinct domains such as inventory metrics and process utilization's. Within these  categories, a single VSM can contain hundreds of data elements, ranging from small JSON objects to long time-series.

\subsection{Datasets}
\label{subsec:data}

Here, we describe the data that was used to develop and evaluate the agent. Due to the unique subject matter of VSM and the specific composition of the data produced in the simulation, a specialized dataset had to be curated. This dataset is structured as functional triples:
\begin{itemize}
    \item \textbf{VSM} (Context): The VSM model and its simulation data for the agent to retrieve information from.
    \item \textbf{Question} (Query): A natural language query formulated by a VSM domain expert to target insights that can be derived from the provided VSM simulation context.
    \item \textbf{Answer} (Ground Truth): The correct answer to a query within a specific VSM simulation context, either authored directly by a VSM expert, or generated by the agent and validated by the VSM expert post-hoc.
\end{itemize}

\label{sec:train-data-eval}
To adhere to the principle of separating development data from test data, we created two datasets containing these triples. The first {\bf development dataset} was created to measure progress during the development of the agent. It contains 12 unique VSM simulation contexts and 13 questions. Each context was used for multiple questions, and vice versa, resulting in 112 unique triples.

Eight of the VSM simulation contexts are industry-inspired, meaning they are modeled after real-world VSMs from customers. These were carefully sanitized and abstracted to protect intellectual property, production secrets, and customer identities. The industries include: Pharmaceuticals, Metalworking, Electronic Production, Water Management and Automotive Manufacturing. The remaining four contexts were generic VSMs designed by the expert to establish a baseline level of complexity and test the fundamental capabilities of the agent.
Examples of the dataset are provided in the appendix~\ref{appendix:dataset-example}.

\label{subsec:eval-data}
To ensure the agent could handle user queries that had not been tested during development, a second {\bf evaluation dataset} was created. To cover higher diversity in tasks, this resulted in an evaluation dataset consisting of 47 triples, containing 20 questions and three contexts. 

\subsection{Agentic Architecture}
\label{subsec:architecture}

In the domain of VSM simulation, extracting meaningful insights requires an agent to dynamically cross-reference static structural parameters with massive, dynamic output logs. Consider the following question: "Are any supermarkets under supplied?" To answer this, an agent cannot simply fetch a single document. It must perform multi-hop reasoning: Firstly it needs to identify the specific "supermarket" nodes within the topology, extract their individual attributes, discover they have "safety\_stock\_limits" attributes for their different products, locate the corresponding dynamic time-series logs for product inventory, and finally evaluate those logs to identify instances where inventory levels dip below the predefined limits.

The proposed architecture accomplishes this by strictly separating the broad logical reasoning process from the data analysis. This is achieved through a decoupled, agentic architecture consisting of a orchestration agent and a summarization subworkflow.
\begin{itemize}
    \item The \textbf{orchestration agent} acts as the high-level planner and reasoning engine. It interprets the user's overarching goal and navigates the simulation data taxonomy to determine where the necessary information resides. Crucially, instead of bloating its own context with raw data and then trying to figure out how to summarize or filter its entire context, the agent never comes into contact with raw simulation data in the first place. This way the orchestration agent maintains a clean context dedicated solely to multi-hop planning and logic. In order for the agent to know what parts of the simulation data are of interest, it is presented with domain expert knowledge, describing the data element and detailing instructions on when which data element is or is not of interest.
    
    \item Once the target data element is identified, the orchestration agent delegates the analysis by triggering a \textbf{summarization subworkflow}. The subworkflow receives the data element and the domain expert knowledge related to it, as well as instructions from the orchestration agent on what insights to extract. It processes the data and returns its summarization back to the orchestration agent. The context is completely refreshed with every call from the orchestration agent, so that unnecessary data does not accumulate.
\end{itemize}

\begin{figure}[!ht]
     \centering
     \includegraphics[scale=0.60]{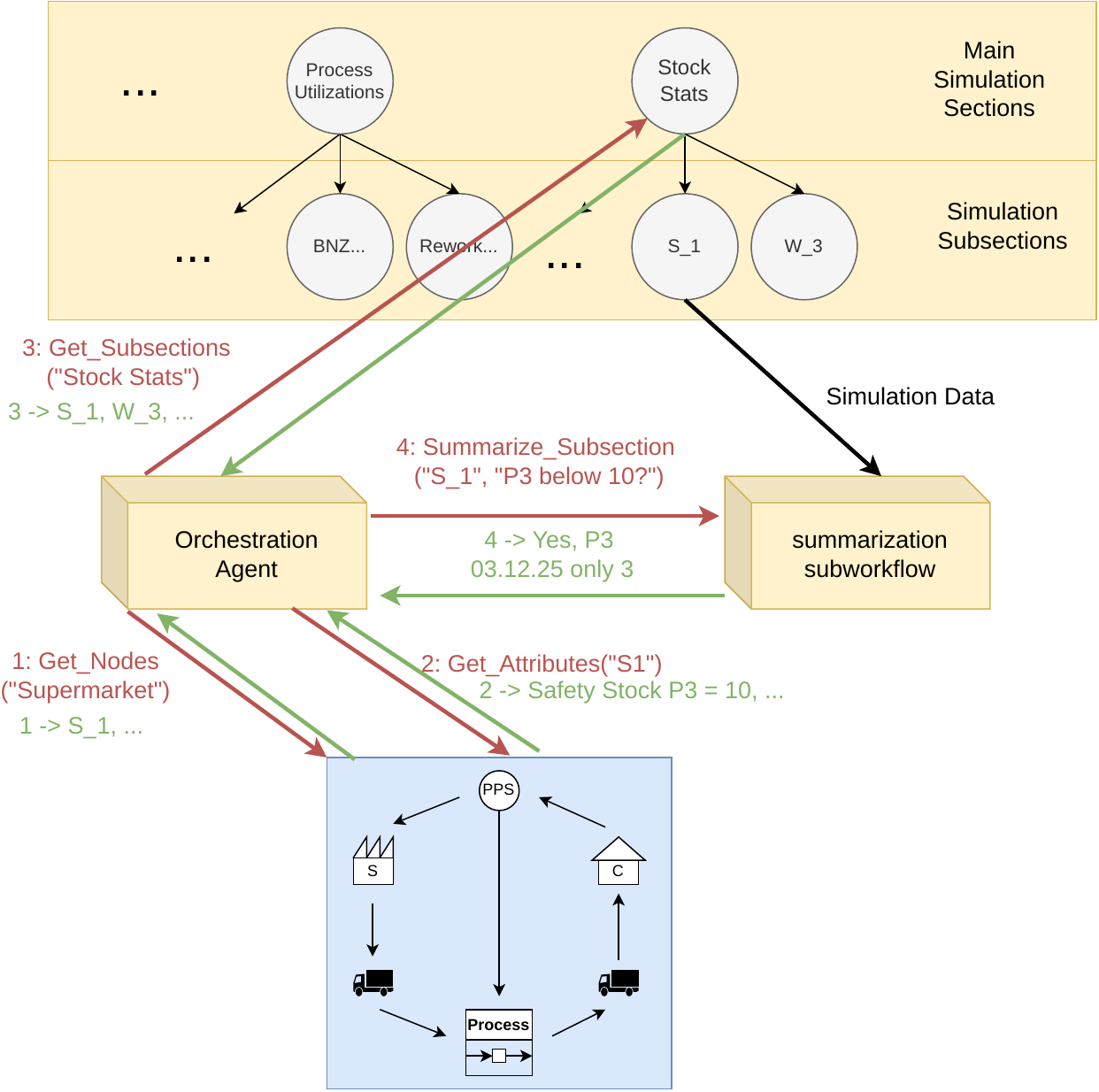}
     \caption{Overview of the decoupled agentic architecture, showcasing four steps of the under-supplied supermarkets example. Red marks tool calls. Green marks the returned information.}
     \label{fig:architecture}
 \end{figure}

This process is repeated until the orchestration agent has gathered all necessary insights to formulate a comprehensive response to the user.
Through this strict separation of concerns, both the orchestration agent and the subworkflow only ever interact with the precise information required for their respective tasks. This design effectively eliminates context rot and enables the system to process massive quantities of diversely formatted data without relying on brittle, hardcoded data processing. Figure~\ref{fig:architecture} showcases four steps of the orchestration agent on the example of earlier ("Are any supermarkets undersupplied?"): 
\begin{itemize}
    \item \textbf{Step 1:} The orchestration agent first finds out which nodes are of type "supermarket". One of the supermarkets is called "S1".
    \item \textbf{Step 2:} It retrieves the attributes of a supermarket node "S1". One of the safety stock levels is 10 for product P3. 
    \item\textbf{Step 3:} It retrieves the data elements of the "Stock Statistic" main simulation section. One of the simulation data elements is called "S\_1".
    \item \textbf{Step 4:} It tasks the summarization subworkflow with checking if the product P3 is ever lower than 10 in the data element "S\_1". The summarization subworkflow analyzes the data element to determine if the stock level of product "P3" ever falls below 10. It finds out that on 03.12.25 the stock level was only 3 and reports its insights back to the orchestration, which formulates a final answer to the user.
\end{itemize}

To execute its reasoning and navigate the dual data structure, the orchestration agent is equipped with a set of four tools. These tools are purposefully designed to enforce progressive discovery, ensuring the agent only accesses information immediately relevant to its current logical step. To interact with the VSM, the agent relies on two complementary tools: 

The {\bf Node Discovery Tool} lists the nodes present within the VSM. It features an optional class-type filter, allowing the agent to request only specific entities (e.g., returning only the names and classes of "supermarkets"). 

The {\bf Attribute Extraction Tool} retrieves the detailed structural attributes (such as capacity or safety stock) of a single, targeted node. Operating in tandem, these tools allow the agent to map the network step-by-step. By first identifying what entities exist and then selectively querying only the necessary parameters, the agent completely avoids the context bloat associated with loading the entire graph.

Aware of the 11 primary simulation sections through its hardcoded tool descriptions, the agent utilizes a {\bf Taxonomy Navigation Tool}. When triggered, this tool returns a list of all available data elements and the domain expert knowledge related them for a specified main section. This enables the orchestration agent to precisely locate data sources within the taxonomy without having to process the descriptions of hundreds of irrelevant data elements. 

The \textbf{summarization Tool} serves as the bridge between structural planning and dynamic analysis. The orchestration agent passes the name of the target data element alongside a customized natural language instruction. The subworkflow receives this instruction, the domain experts knowledge related to the data element, and the data itself. It then performs its isolated analysis and returns it to the orchestration agent.

\section{EVALUATION}
\label{sec:eval}

In our experiments, we assess the efficacy and stability of the proposed decoupled agentic architecture. Section~\ref{subsec:setup} establishes the experimental setup, detailing the LLMs, hardware, software frameworks and sampling parameters used in this work. Following this, Section~\ref{subsec:llm-as-judge} outlines and validates the automated evaluation methodology used to measure the agent's performance. The core empirical results are presented in Section~\ref{subsec:quantitative-eval} followed by a qualitative analysis identifying primary causes of failures in Section~\ref{subsec:qualitative-eval}. Finally, acknowledging the significant computational overhead associated with LLM inference, Section~\ref{sec:carbon} estimates the energy consumption of this study and details the measures taken to offset the resulting carbon emissions.

\subsection{Setup}\label{subsec:setup}

To assess the robustness and scalability of the proposed approach, it was evaluated using different LLM configurations.
The models were selected based on their performance relative to their size. They span from efficient open-source to SOTA closed-source LLMs and cover different architectures and publishers. Based on that the following LLMs were chosen: 
\begin{itemize}
    \item \textbf{Ministral-3-14B-Reasoning-2512}~\citep{liu2026ministral} an edge-optimized dense model
    \item \textbf{Qwen3-30B-A3B-Thinking-2507}~\citep{yang2025qwen3} an efficient Mixture-of-Experts (MoE)
    \item \textbf{GLM-5}~\citep{zeng2026glm} a 744B parameters (40B active) MoE
    \item \textbf{Claude-Opus 4.6}~\citep{anthropic2026claudeopus46} a SOTA closed source model
\end{itemize}
Ministral and Qwen were hosted on two NVIDIA RTX A6000s via vLLM~\citep{kwon2023efficient}. The larger LLMs were accessed through their respective publishers' APIs.

The proposed architecture was implemented in DSPy~\citep{khattab2022demonstrate}, with its parameters left at their default configuration, except for \textbf{max\_retries}, which was set to 5. This parameter controls how many times a model is prompted again if it fails to produce structured output.

For the agent, a \textbf{temperature}~\citep{ACKLEY1985147} of 0.3 was used across all evaluations. 

\textbf{max\_tokens} so the maximum amount of tokens that a single LLM response is allowed to contain was set to 20k. This amount was sufficient for even the longest data-series present in the dataset. Capping it at 20k stopped generation loops earlier, thereby speeding up the evaluation process and saving resources. This is particularly important given that many LLM APIs incur additional costs for contexts larger than this amount.

Because LLMs are inherently susceptible to prompt sensitivity~\citep{zhuo-etal-2024-prosa} and exhibit non-deterministic outputs, relying on a single evaluation can yield skewed results. To counteract this statistical randomness, all conducted \textbf{evaluations were performed four times}.

\subsection{LLM as a judge}
\label{subsec:llm-as-judge}

We follow a quantitative evaluation protocol, in which the triples (context = VSM + simulation data, question, answer) are run through the agent. The correctness of the agents response is rated as floating-point value between 0 for wrong responses and 1 for correct ones. This is done by a LLM as a judge method.

An evaluation was conducted to determine the most suitable LLM for judging the following experiments, and to validate the evaluation protocol as a whole. Thirty triples from the evaluation dataset were processed by an agent.
The responses were then evaluated by two VSM experts and multiple LLMs. The two VSM experts had access to the VSM simulation context, the correct answer, the query, and the agent's response. To avoid influencing their ratings, the experts were not shown each other's ratings or those of the LLM judges. The ratings of the four evaluation runs of the LLMs judges were aggregated and processed to compare model performance against VSM expert judgments. This was done in three stages:
\begin{enumerate}
    \item \textbf{Establishing the Human Baseline Agreement rate:} Before evaluating the models, we established an empirical upper bound for agreement and overall rating by measuring the ratings of two human experts. We calculated the overall mean rating, the mean absolute error ($\text{MAE} = \frac{1}{n} \sum_{i=1}^{n} |x_i - y_i|$), and the Pearson correlation coefficient ($r$):
    $$r = \frac{\sum (x_i - \bar{x})(y_i - \bar{y})}{\sqrt{\sum (x_i - \bar{x})^2 \sum (y_i - \bar{y})^2}}$$
    between the ratings of Expert 1 and Expert 2. This baseline represents the realistic maximum agreement that can be expected from an LLM, accounting for natural human subjectivity. 
    \item \textbf{Formulating the Ground Truth:} To create a single, stable target against which to evaluate the models, a consensus "ground truth" rating was calculated for each sample. This was achieved by averaging the two VSM expert ratings.
    \item \textbf{Evaluating Model Performance:} Finally, the ratings generated by each respective LLM judge were compared against the averaged human ground truth. For each model, we computed the MAE to measure the average absolute deviation from the consensus score, the Pearson correlation to assess how well the model aligned with the human experts' relative ranking of the samples, and the overall mean rating to determine if one judge was harsher than the other.
\end{enumerate}

The results of the evaluation can be seen in Table~\ref{tab:judge-eval}.
\begin{table}[h!]
    \centering
    \caption{LLM as a judge validation performed on a sub collection of the evaluation dataset. Ratings were multiplied by 100 to set their interval between 0 and 100. Ratings of the for evaluation runs of LLM judges were averaged. Mean rating should be as close as possible to the human base line. MAE  should be as low as possible. Pearson r should be as high as possible. Bold marks best LLM metric.}
    \begin{tabular}{c|c|c|c}
         Evaluator  & Mean Rating & MAE $\downarrow$ & Pearson r $\uparrow$ \\ \hline
         Expert Consensus & 76.67 & 10.67 & 80.70  \\ \hline
         Qwen3-30B-A3B-Thinking-2507 & 72.67 & 15.67 & 69.92 \\
         GPT-4.1 & 73.33 & \textbf{7.33} & \textbf{89.99} \\
         GPT-5-mini & \textbf{77.33} & 11 & 78.51 \\
         GPT-5.4 & 67.33 & 12.67 & 79.77 \\
    \end{tabular}
    \label{tab:judge-eval}
\end{table}

The expert agreement baseline established a robust ground truth for the evaluation, exhibiting a mean rating of 76.67, an average human-to-human deviation (MAE) of 10.67, and a Pearson correlation of 80.70 between the two experts. Among the evaluated LLM judges, GPT-4.1 emerged as the strongest overall candidate. It recorded the lowest MAE (7.33) and the highest Pearson correlation (89.99). While these metrics numerically surpass the inter-expert baseline, this is an statistical artifact of evaluating the models against a smoothed consensus average rather than the noisier variance between two individual human raters. Notably, OpenAI API constraints currently mandate a temperature of 1 for the GPT-5 models, whereas GPT-4.1 and Qwen3 were evaluated at a deterministic temperature of 0; this forced randomness likely explains why the GPT-5 models were outperformed here. Based on these strong agreement metrics, GPT-4.1 was selected as the primary judge for all subsequent evaluations. Because GPT-4.1 exhibits a slight strictness bias (mean 73.33 vs. human consensus 76.67), the downstream results presented in this study can be interpreted as a conservative lower bound of what human experts would likely assign.

\subsection{Quantitative Cross-Model Benchmarking}
\label{subsec:quantitative-eval}

The quantitative evaluation of the agent under each model configuration, as detailed in Section~\ref{subsec:setup}, was measured on the evaluation dataset established in Section~\ref{subsec:eval-data}, utilizing the LLM-as-a-judge method from Section~\ref{subsec:llm-as-judge} is shown in Table~\ref{tab:agent-eval}.

\begin{table}[h!]
    \centering
    \caption{Agent evaluation across 4 evaluation runs on the full evaluation dataset. The score in rating row represent the average rating across all samples followed by the standard divination (SD) across evaluation runs. Sample Token AVG denotes the average amount of tokens used to process a single triple.}
    \begin{tabular}{c|p{2.9cm}p{2.9cm}cc}
         Model& Ministral-3-14B-Reasoning-2512 & Qwen3-30B-A3B-Thinking-2507 & GLM-5 & Claude-Opus 4.6\\ \hline
         AVG ± SD Rating & 57.34 ± 6.64 & 72.66 ± 2.1 & 83.88 ± 3.92 & 85.96 ± 0.86\\ \hline
         Sample Token AVG & 39,819 & 43,482 & 27,580 & 38,594\\
    \end{tabular}
    \label{tab:agent-eval}
\end{table}

The evaluation results reveal a performance hierarchy that correlates with architectural scale. Claude-Opus 4.6, representing a SOTA closed-source frontier model, achieved the highest overall rating (85.96). However, the massive open-weights GLM-5 closely trailed it with a score of 83.88. This observation strongly aligns with recent literature demonstrating that top-tier open-weight models are closing the performance gap with proprietary systems~\citep{grattafiori2024llama,manchanda2024open}. Conversely, the explicitly scaled-down models—Qwen-30B-A3B and Ministral-14B—scored predictably lower at 72.66 and 57.34, respectively. This confirms that while smaller open-source models are highly accessible, raw parameter count and vast pre-training remain heavily influential in absolute reasoning quality.

An analysis of the SD yields a notable inverse relationship between model size or performance and output variance. The smallest model, Ministral-14B, struggled significantly with consistency, exhibiting a massive SD of 6.64 across the four runs. This high variance also suggests that the model itself is uncertain about its responses~\citep{KuhnGF23}. As model size and capability increase, the variance heavily compresses. The mid-sized Qwen-30B stabilized to a SD of 2.10, and while GLM-5 exhibited a slight fluctuation with a SD of 3.92, it still maintained significantly better consistency than the baseline model. Ultimately, the leading Claude-Opus 4.6 proved robust, with a near-zero SD of 0.86. This highlights that frontier models not only produce better answers, but also exhibit greater reliability.

Finally, it is worth noting that performance does not strictly correlate with token verbosity. GLM-5 achieved near-peak performance while using the lowest average tokens per sample (27,580), demonstrating efficient context utilization and direct reasoning compared to the more verbose Qwen (43,482 tokens).

\subsection{Qualitative Observations}
\label{subsec:qualitative-eval}

A qualitative analysis of the failure cases across all LLM configurations reveals a consistent struggle with specific metrics, such as process rework amounts. Around 15\% of samples were never answered correctly by any LLM configuration. The information needed to resolve the majority of these cases did not require multi-hop reasoning or large-scale data summarization but could simply be retrieved from a single simulation data element. Rather than arbitrary hallucinations, the models systematically retrieved incorrect values from wrong data sources. This indicates that the primary failure point is not the summarization of the subworkflow, but rather the navigation of the orchestration agent, likely due to  misleading or semantically ambiguous domain expert knowledge in the simulation data topology.

\subsection{Carbon Offsetting} \label{sec:carbon}
CO$_2$ emissions estimations were conducted using the Machine Learning Impact calculator presented in~\citep{lacoste2019quantifying}. For the locally hosted ministral and qwen 10.37 kg CO$_2$eq was estimated. For models used via cloud APIs, the exact hardware, utilization time, and specific geographic routing are obscured. Based on model sizes (partly estimated), region (also estimated) and runtime, 87.81 kg CO$_2$eq was calculated.
To offset the consumed energy, the technology focused plan of~\citep{climeworks} was used to remove 100 kg CO$_2$.

\section{CONCLUSION}
\label{sec:conc}

In this paper, we present a decoupled, agent-based architecture designed to extract valuable insights from complex VSM simulations.
The system uses an orchestrator agent for progressive data discovery, infused domain expert knowledge and a specialized analysis subworkflow. This enables the system to bridge the gap between structural simulation model parameters and dynamic simulation data while effectively mitigating context rot. Empirical evaluation within the VSM simulation domain demonstrated the framework's high efficacy. SOTA LLMs achieved high accuracy by scoring up to ~86/100 and demonstrated great stability across evaluation runs.

Despite these strong results, the presented approach faces multiple limitations. 
The progressive data discovery worked unreliably with current small scale LLMs. Additionally, relying exclusively on native LLM reasoning to evaluate large amounts of data is computationally intensive, leading to high token consumption and increased latency. Furthermore, this simple approach is highly dependent on the model being able to process whole time-series at once. While there are a wide range of workarounds to this issue, and the tested LLMs rarely struggled with data analysis during evaluation, our data did not include longer time series that exceeded the context window of the LLM.

To address these limitations, future work can focus on several key areas. First, other agentic architectures might be able to further simplify the progressive data discovery, enabling especially smaller LLMs to perform the task more reliably. Second, to scale the analysis for larger data i.e. time-series a more complex summarization subworkflow could be employed. Data pre-processing might reduce token consumption and thereby enable the subworkflow to process data efficiently. Finally, future research should explore closing the simulation based optimization loop by extending the architecture into one that is able to provide well-founded optimization suggestions by intelligently modifying simulation parameters, trigger new simulations and evaluate produced simulation data all autonomously.

In conclusion, LLM-based agents show promising results for analyzing complex VSM simulation environments. However, they still fall short on reliably providing domain expert level insights.

\section*{ACKNOWLEDGMENTS}
This work was funded by the German Federal Ministry of Research, Technology and Space (Project "VaStNet", ID 01IS24015C).

\appendix

\section{DATASET EXAMPLES}\label{appendix:dataset-example}

\begin{table}[h!]
    \centering
    \caption{From German to English translated data examples used in this work.}
    \begin{tabular}{c|p{5cm}|p{8cm}}
        VSM name & question & answer \\ \hline
        pharma & Which process is the bottleneck in the overall system? & The process bottling is with a utilization rate of 59.75\% the bottleneck in the overall system.\\ \hline
        automotive-01 & Which process has the highest overall throughput? & The Outer\_Stamping station has the highest total throughput with 27094 pieces. \\ \hline
        spray-02 & How long and how often does a storage tank/buffer run empty during the simulation time? (referring to the minimum storage level in parts) & The following warehouses/supermarkets ran out of stock during the simulation (10 days):
        Warehouse, Warehouse\_02, and Warehouse\_03 ran empty from the start and remained so. Warehouse\_01 was filled on March 4, 2021, and ran empty again from March 8, 2021. 
        Warehouse\_04 Component Base\_Lid\_Skids ran out of stock on March 6, 2021, and then gradually refilled.
        Component KSWg ran empty from March 4, 2021, for the remainder of the simulation. \\
    \end{tabular}
\end{table}

\footnotesize

\bibliographystyle{plainnat}
\bibliography{agentic}

\end{document}